\documentclass[11pt,a4paper]{article}
\usepackage[hyperref]{emnlp2019}
\usepackage{times}
\usepackage{latexsym}

\usepackage{tikz}
\definecolor{forestgreen}{rgb}{0.0, 0.27, 0.13}
\definecolor{burntorange}{rgb}{0.8, 0.33, 0.0}

\usepackage{pifont}

\usepackage[normalem]{ulem}

\usepackage[utf8]{inputenc}
\usepackage{color,soul}
\setul{0.45ex}{0.25ex}

\newcommand\boldcolor[2]{\textcolor{#1}{\textbf{#2}}}
\newcommand\answer[2]{\boldcolor{#1}{\setulcolor{#1}\ul{#2}}}

\usepackage{makecell}
\usepackage{graphicx}
\usepackage{subcaption}
\usepackage{booktabs}
\usepackage{multirow}
\usepackage{graphicx}
\usepackage{array}
\usepackage{pgfplots}
\usepackage{amsmath}
\usepackage{algorithm}
\usepackage{url}

\usepackage{xcolor}
\usepackage{times}
\usepackage{latexsym}
\usepackage{amsmath}
\usepackage{algorithm}
\usepackage{amsfonts}
\usepackage[noend]{algpseudocode}
\usepackage{lipsum}
\usepackage{multirow}
\usepackage{framed}
\usepackage{graphicx}
\usepackage{longtable}
\usepackage{tikz}
\usepackage{booktabs}
\usepackage[all]{nowidow}
\usepackage{soul}
\usepackage{enumitem}
\definecolor{coloranswer}{HTML}{0596FF}
\definecolor{lightblue}{HTML}{3CC7EA}
\definecolor{boxcolor}{HTML}{EEEEEE}
\definecolor{periwinkle}{rgb}{0.8, 0.8, 1.0}
\definecolor{colorsquad}{rgb}{0,1,0}
\definecolor{colorsnli}{rgb}{1,0,0} 
\definecolor{colorvqa}{rgb}{1,1,0}

\aclfinalcopy 

\let\oldemptyset\emptyset
\let\emptyset\varnothing

\newcommand{\qanet}{NAQANet}
\newcommand{\drop}{DROP}
\newcommand{\glove}{GloVe}
\newcommand{\wtovec}{word2vec}
\newcommand{\equate}{EQUATE}
\newcommand{\elmo}{ELMo}
\newcommand{\bert}{BERT}              
\newcommand{\bertbase}{BERT-base}

\newcommand*\samethanks[1][\value{footnote}]{\footnotemark[#1]}
\usepackage{amssymb}

\title{Do NLP Models Know Numbers? Probing Numeracy in Embeddings}

\author{\makecell{Eric Wallace\thanks{~~Equal contribution; work done while interning at AI2.}$~~^1$, Yizhong Wang\samethanks$~~^2$, Sujian Li$^2$, Sameer Singh$^3$, Matt Gardner$^1$}\\
$^1$Allen Institute for Artificial Intelligence\\
$^2$Peking University\\ 
$^3$University of California, Irvine\\
\{\href{mailto:ericw@allenai.org}{\tt ericw}, \href{mailto:mattg@allenai.org}{\tt mattg}\}\href{mailto:mattg@allenai.org}{\tt @allenai.org},  
\{\href{mailto:yizhong@pku.edu.cn}{\tt yizhong}, \href{mailto:lisujian@pku.edu.cn}{\tt lisujian}\}\href{mailto:lisujian@pku.edu.cn}{\tt @pku.edu.cn},
\href{mailto:sameer@uci.edu}{\tt sameer@uci.edu}}

\date{}

\begin{document}
\maketitle
\begin{abstract}
The ability to understand and work with numbers (numeracy) is critical for many complex reasoning tasks. Currently, most NLP models treat numbers in text in the same way as other tokens---they embed them as distributed vectors. Is this enough to capture numeracy? We begin by investigating the numerical reasoning capabilities of a state-of-the-art question answering model on the DROP dataset. We find this model excels on questions that require numerical reasoning, i.e., it already captures numeracy. To understand how this capability emerges, we probe token embedding methods (e.g., BERT, GloVe) on synthetic list maximum, number decoding, and addition tasks. A surprising degree of numeracy is naturally present in standard embeddings. For example, GloVe and word2vec accurately encode magnitude for numbers up to 1,000. Furthermore, character-level embeddings are even more precise---ELMo captures numeracy the best for all pre-trained methods---but BERT, which uses sub-word units, is less exact.
\end{abstract}
  
\section{Introduction}\label{sec:intro}

\begin{figure}[t] 
\centering
\includegraphics[trim={2.6cm 8cm 2.7cm 7.7cm},clip, width=\columnwidth]{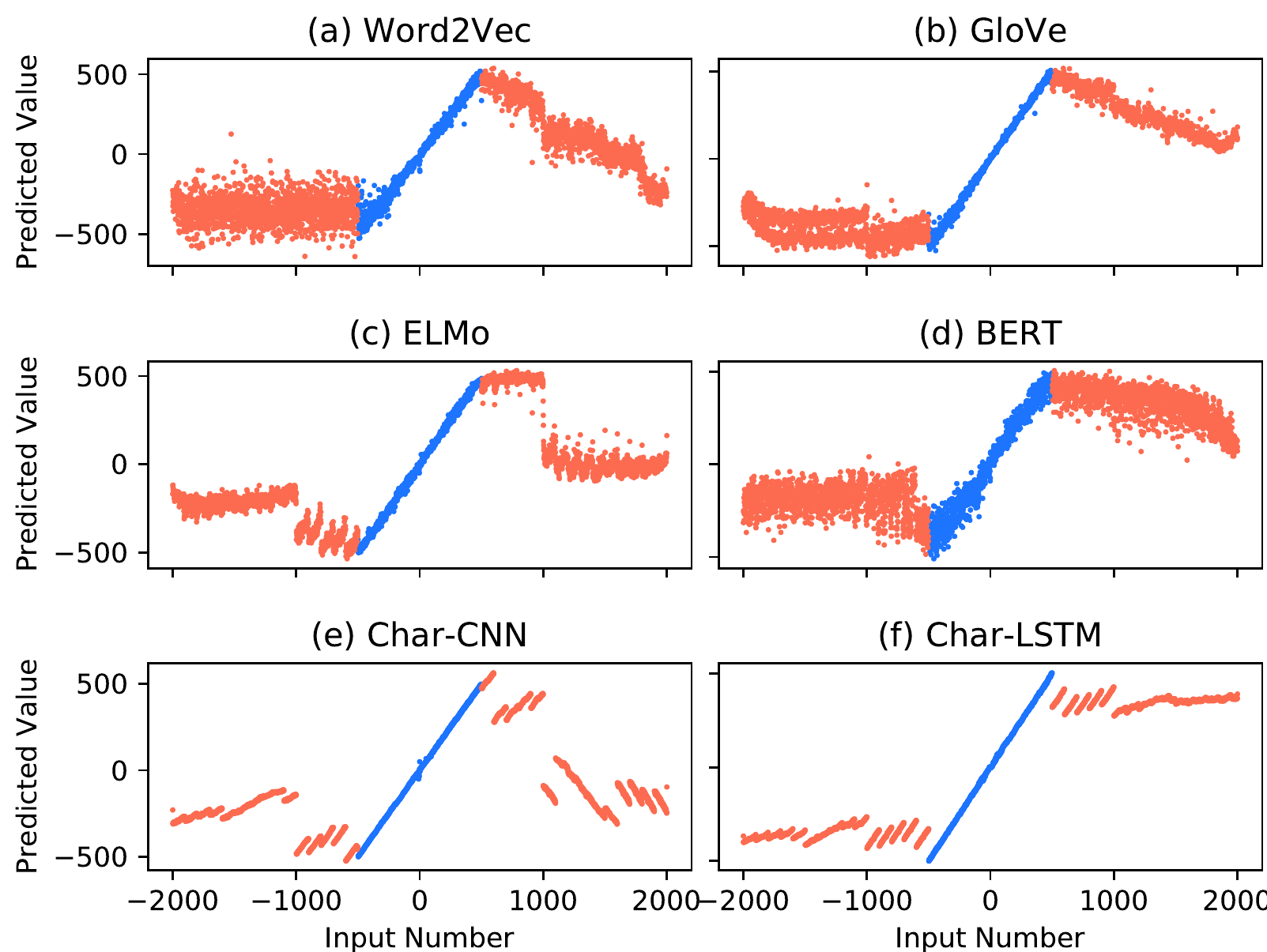}
  \caption{We train a probing model to decode a number from its word embedding over a random 80\% of the integers from [-500, 500], e.g., ``71'' $\to$ 71.0. We plot the model's predictions for all numbers from [-2000, 2000]. The model accurately decodes numbers within the training range (in \textcolor{blue}{blue}), i.e., pre-trained embeddings like GloVe and BERT capture numeracy. However, the probe fails to extrapolate to larger numbers (in \textcolor{red}{red}). The Char-CNN (e) and Char-LSTM (f)  are trained jointly with the probing model.}
  \label{fig:plot} 
\end{figure}

Neural NLP models have become the de-facto standard tool across language understanding tasks, even solving basic reading comprehension and textual entailment datasets~\cite{yu2018qanet,devlin2018BERT}. Despite this, existing models are incapable of complex forms of reasoning, in particular, we focus on the ability to reason \emph{numerically}. Recent datasets such as \drop{}~\cite{drop}, \equate{}~\cite{ravichander2019equate}, or Mathematics Questions~\cite{saxton2019analysing} test numerical reasoning; they contain examples which require comparing, sorting, and adding numbers in natural language (e.g., Figure~\ref{fig:example_1}).

The first step in performing numerical reasoning over natural language is \emph{numeracy}: the ability to understand and work with numbers in either digit or word form~\cite{Spithourakis2018numeracy}. For example, one must understand that the string ``23'' represents a bigger value than ``twenty-two''.  Once a number's value is (perhaps implicitly) represented, reasoning algorithms can then process the text, e.g., extracting the list of field goals and computing that list's maximum (first question in Figure~\ref{fig:example_1}). Learning to reason numerically over paragraphs with only question-answer supervision appears daunting for end-to-end models; our work seeks to understand \emph{if} and \emph{how} ``out-of-the-box'' neural NLP models already learn this. 

We begin by analyzing the state-of-the-art \qanet{} model \citep{drop} for \drop{}---testing it on a subset of questions that evaluate numerical reasoning (Section~\ref{sec:drop}). To our surprise, the model exhibits excellent numerical reasoning abilities. Amidst reading and comprehending natural language, the model successfully computes list maximums/minimums, extracts superlative entities (\texttt{argmax} reasoning), and compares numerical quantities. For instance, despite \qanet{} achieving only 49 F1 on the entire validation set, it scores 89 F1 on numerical comparison questions. We also stress test the model by perturbing the validation paragraphs and find one failure mode: the model struggles to extrapolate to numbers outside its training range.

We are especially intrigued by the model's ability to learn numeracy, i.e., how does the model know the value of a number given its embedding? The model uses standard embeddings (\glove{} and a Char-CNN) and receives no direct supervision for number magnitude/ordering. To understand how numeracy emerges, we probe token embedding methods (e.g., BERT, GloVe) using synthetic list maximum, number decoding, and addition tasks (Section~\ref{sec:numeracy}). 

We find that all widely-used pre-trained embeddings, e.g.,  \elmo{}~\cite{PetersELMo2018}, \bert{}~\cite{devlin2018BERT}, and \glove{}~\cite{pennington2014glove}, capture numeracy: number magnitude is present in the embeddings, even for numbers in the thousands. Among all embeddings, character-level methods exhibit stronger numeracy than word- and sub-word-level methods (e.g., \elmo{} excels while \bert{} struggles), and character-level models learned directly on the synthetic tasks are the strongest overall. Finally, we investigate why \qanet{} had trouble extrapolating---was it a failure in the model or the embeddings? We repeat our probing tasks and test for model extrapolation, finding that neural models struggle to predict numbers outside the training range.
\section{Numeracy Case Study: DROP QA}\label{sec:drop}

\begin{figure}[t]
\centering
\small
\tikz\node[draw=black!40,inner sep=1pt,line width=0.3mm,rounded corners=0.1cm]{
\begin{tabular}{p{0.46\textwidth}}
\dots JaMarcus Russell completed a \setulcolor{burntorange}\ul{91-yard} touchdown pass to rookie wide receiver \answer{magenta}{Chaz Schilens}.  The Texans would respond with fullback Vonta Leach getting a {1-yard touchdown run}, yet the Raiders would answer with kicker Sebastian Janikowski getting a \setulcolor{cyan}\ul{33}-yard and a \setulcolor{cyan}\ul{21}-yard field goal.  Houston would tie the game in the second quarter with kicker Kris Brown getting a \answer{cyan}{53}-yard and a \setulcolor{cyan}\ul{24}-yard field goal. Oakland would take the lead in the third quarter with wide receiver \setulcolor{magenta}\ul{Johnnie Lee Higgins} catching a \answer{burntorange}{29-yard} touchdown pass from Russell, followed up by an 80-yard punt return for a touchdown.
\end{tabular}
};

\vspace{0.1cm}
\tikz\node[draw=black!20!cyan,inner sep=1pt,line width=0.3mm,rounded corners=0.1cm]{
\begin{tabular}{p{0.46\textwidth}}
Q: How many yards was the longest field goal? A: 53
\end{tabular}
};

\vspace{0.1cm}
\tikz\node[draw=burntorange,inner sep=1pt,line width=0.3mm,rounded corners=0.1cm]{
\begin{tabular}{p{0.46\textwidth}}
Q: How long was the shortest touchdown pass? A: 29-yard
\end{tabular}
};

\vspace{0.1cm}
\tikz\node[draw=black!20!magenta,inner sep=1pt,line width=0.3mm,rounded corners=0.1cm]{
\begin{tabular}{p{0.46\textwidth}}
Q: Who caught the longest touchdown? A: Chaz Schilens
\end{tabular}
};
\caption{Three \drop{} questions that require numerical reasoning; the state-of-the-art \qanet{} answers every question correct. Plausible answer candidates to the questions are underlined and the model's predictions are shown in bold.}
\label{fig:example_1} 
\end{figure}

This section examines the state-of-the-art model for \drop{} by investigating its accuracy on questions that require numerical reasoning.

\subsection{DROP Dataset}\label{subsec:drop}
	
\drop{} is a reading comprehension dataset that tests numerical reasoning operations such as counting, sorting, and addition~\cite{drop}. The dataset's input-output format is a superset of SQuAD~\cite{rajpurkar2016squad}: the answers are paragraph spans, as well as question spans, number answers (e.g., 35), and dates (e.g., 03/01/2014).  The only supervision provided is the question-answer pairs, i.e., a model must learn to reason numerically while simultaneously learning to read and comprehend.

\subsection{\qanet{} Model}

Modeling approaches for \drop{} include both semantic parsing~\cite{krishnamurthy2017semantic} and reading comprehension~\cite{yu2018qanet} models. We focus on the latter, specifically on Numerically-augmented QANet (\qanet{}),  the current state-of-the-art model~\cite{drop}.\footnote{Result as of May 21st, 2019.} The model's core structure closely follows QANet~\cite{yu2018qanet} except that it contains four output branches, one for each of the four answer types (passage span, question span, count answer, or addition/subtraction of numbers.) 

Words and numbers are represented as the concatenation of \glove{} embeddings and the output of a character-level CNN. The model contains no auxiliary components for representing number magnitude or performing explicit comparisons. We refer readers to \citet{yu2018qanet} and \citet{drop} for further details.

\subsection{Comparative and Superlative Questions}

\begin{table*}[t]
\centering
\small
\begin{tabular}{lll}
\toprule
\bf Question Type & \bf Example  & \bf Reasoning Required   \\ \midrule
Comparative (Binary) & Which country is a bigger exporter, Brazil or Uruguay? & Binary Comparison \\
Comparative (Non-binary) & Which player had a touchdown longer than 20 yards? & Greater Than \\
Superlative (Number) & How many yards was the shortest field goal? & List Minimum \\
Superlative (Span) & Who kicked the longest field goal? & Argmax \\
\bottomrule
\end{tabular}
\caption{We focus on \drop{} \emph{Comparative} and \emph{Superlative} questions which test \qanet{}'s numeracy.}
\label{table:examples}
\end{table*}

We focus on questions that \qanet{} requires numeracy to answer, namely \emph{Comparative} and \emph{Superlative} questions.\footnote{\drop{} addition, subtraction, and count questions do not require numeracy for \qanet{}, see Appendix~\ref{appendix:qanet}.} 
Comparative questions probe a model's understanding of quantities or events that are ``larger'', ``smaller'', or ``longer'' than others. Certain comparative questions ask about ``either-or'' relations (e.g., first row of Table~\ref{table:examples}), which test binary comparison. Other comparative questions require more diverse comparative reasoning, such as greater than relationships (e.g., second row of Table~\ref{table:examples}). 

Superlative questions ask about the ``shortest'', ``largest'', or ``biggest'' quantity in a passage. When the answer type is a \emph{number}, superlative questions require finding the maximum or minimum of a list (e.g., third row of Table~\ref{table:examples}). When the answer type is a \emph{span}, superlative questions usually require an \texttt{argmax} operation, i.e., one must find the superlative action or quantity and then extract the associated entity (e.g., fourth row of Table~\ref{table:examples}). We filter the validation set to comparative and superlative questions by writing templates to match words in the question. 

\subsection{Emergent Numeracy in \qanet{}}

\qanet{}'s accuracy on comparative and superlative questions is significantly higher than its average accuracy on the validation set (Table~\ref{table:breakdown}).\footnote{We have a public \qanet{} demo available \url{https://demo.allennlp.org/reading-comprehension}.}

\qanet{} achieves 89.0 F1 on binary (either-or) comparative questions, approximately 40 F1 points higher than the average validation question and within 7 F1 points of human test performance. The model achieves a lower, but respectable, accuracy on non-binary comparisons. These questions require multiple reasoning steps, e.g., the second question in Table~\ref{table:examples} requires (1) extracting all the touchdown distances, (2) finding the distance that is greater than twenty, and (3) selecting the player associated with the touchdown of that distance. 

We divide the superlative questions into questions that have number answers
and questions with span answers according to the dataset's provided answer type. 
\qanet{} achieves nearly 70 F1 on superlative questions with number answers, i.e., it can compute list maximum and minimums. The model answers about two-thirds of superlative questions with span answers correctly (66.3 F1), i.e., it can perform \texttt{argmax} reasoning.

\begin{table}[t]
\centering \small
\begin{tabular}{lccc}
 \toprule
\textbf{Question Type} & \textbf{Count} & \textbf{EM} & \textbf{F1} \\
\midrule
Human (Test Set) & 9622 & 92.4 & 96.0 \\
\addlinespace
Full Validation & 9536 & 46.2 & 49.2 \\
\hspace{0.5cm} Number Answers & 5842 & 44.3 & 44.4 \\
\addlinespace
Comparative & 704 & 73.6 & 76.4 \\
\hspace{0.5cm} Binary (either-or) & 477 & 86.0 & 89.0 \\
\hspace{0.5cm} Non-binary & 227 & 47.6 & 49.8 \\
\addlinespace
Superlative Questions & 861 & 64.6 & 67.7 \\
\hspace{0.5cm} Number Answers & 475 & 68.8 & 69.2 \\
\hspace{0.5cm} Span Answers & 380 & 59.7 & 66.3 \\
\bottomrule
\end{tabular}
\caption{\qanet{} achieves higher accuracy on questions that require numerical reasoning (\emph{Superlative} and \emph{Comparative}) than on standard validation questions. Human performance is reported from \citet{drop}.}
\label{table:breakdown}
\end{table}

\begin{table}[tb]
\centering
\small
\begin{tabular}{lcccc}
\toprule
\multirow{2}{*}{\textbf{Stress Test Dataset}} & \multicolumn{2}{c}{\textbf{All Questions}}                & \multicolumn{2}{c}{\textbf{Superlative}}        \\ 
                                      & F1 & $\Delta$ &  F1 & $\Delta$ \\ 
\cmidrule(r){1-1}
\cmidrule(lr){2-3}
\cmidrule(lr){4-5}
Original Validation Set & 49.2 & - & 67.7 & -                 \\ 
\addlinespace
{Add {[}1, 20{]}} & 47.7 & -1.5 & 64.1 & -3.6                  \\
{Add {[}21, 100{]} }                    & 41.4                  & -7.8                   & 40.4                  & -27.3                  \\ 
\addlinespace
{Multiply {[}2, 10{]}}                  & 41.1                  & -8.1                   & 39.3                  & -28.4                  \\
{Multiply {[}11, 100{]}}                & 38.8                  & -10.4                   & 32.0                  & -35.7                  \\ 
\addlinespace
{Digits to Words {[}0, 20{]}}           & 45.5                  & -3.7                   & 63.8                  & -3.9                  \\
{Digits to Words {[}21, 100{]}}         & 41.9                  & -7.3                   & 46.1                  & -21.6                  \\ 
\bottomrule
\end{tabular}
\caption{We stress test \qanet{}'s numeracy by manipulating the numbers in the validation paragraphs. \emph{Add} or \emph{Multiply [x, y]} indicates adding or multiplying all of the numbers in the passage by a random integer in the range [x, y]. \emph{Digits $\to$ Words [x, y]} converts all integers in the passage within the range [x, y] to their corresponding word form (e.g., ``75'' $\to$ ``seventy-five'').}
\label{table:stress_test}
\end{table}

Figure~\ref{fig:example_1} shows examples of superlative questions answered correctly by \qanet{}. The first two questions require computing the maximum/minimum of a list: the model must recognize which digits correspond to field goals and touchdowns passes, and then extract the maximum/minimum of the correct list. The third question requires \texttt{argmax} reasoning: the model must first compute the longest touchdown pass and then find the corresponding receiver ``Chaz Schilens''.

\subsection{Stress Testing \qanet{}'s Numeracy}\label{subsec:limit}

Just how far does the numeracy of \qanet{} go? Here, we stress test the model by automatically modifying \drop{} validation paragraphs.

We test two phenomena: larger numbers and word-form numbers. For larger numbers,
we generate a random positive integer and multiply or add that value to the numbers in each paragraph. For word forms, we replace every digit in the paragraph with its word form (e.g., ``75'' $\to$ ``seventy-five''). Since word-form numbers are usually small in magnitude when they occur in \drop{}, we perform word replacements for integers in the range [0, 100]. We guarantee the ground-truth answer is still valid by only modifying \qanet{}'s internal representation (Appendix~\ref{appendix:internal}).

Table~\ref{table:stress_test} shows the results for different paragraph modifications. The model exhibits a tiny degradation in performance for small magnitude changes (e.g., \qanet{} drops 1.5 F1 overall for Add [1,20]) but severely struggles on larger changes (e.g., \qanet{} drops 35.7 F1 on superlative questions for Multiply [11,200]). 
Similar trends hold for word forms: the model exhibits small drops in accuracy when converting small numbers to words (3.9 degradation on Digits to Words [0,20]) but fails on larger magnitude word forms (21.6 F1 drop over [21,100]). These results show that \qanet{} has a strong understanding of numeracy for numbers in the training range, but, the model can fail to extrapolate to other values.

\subsection{Whence this behavior?}
\qanet{} exhibits numerical reasoning capabilities that exceed our expectations. What enables this behavior? Aside from reading and comprehending the passage/question, this kind of numerical reasoning requires two components: numeracy (i.e., representing numbers) and comparison algorithms (i.e., computing the maximum of a list). 

Although the natural emergence of comparison algorithms is surprising, previous results show neural models are capable of learning to count and sort synthetic lists of scalar values when given explicit supervision~\cite{weiss2018counting,vinyals2016set}. \qanet{} demonstrates that a model can learn comparison algorithms while simultaneously learning to read and comprehend, even with only question-answer supervision.

How, then, does \qanet{} know numeracy? The source of numerical information eventually lies in the token embeddings themselves, i.e.,  the character-level convolutions and \glove{} embeddings of the \qanet{} model. Therefore, we can understand the source of numeracy by isolating and probing these embeddings.
\begin{figure*}[tbh] 
\centering
\includegraphics[trim={0cm 1cm 2cm 1.5cm},clip, width=0.8\textwidth]{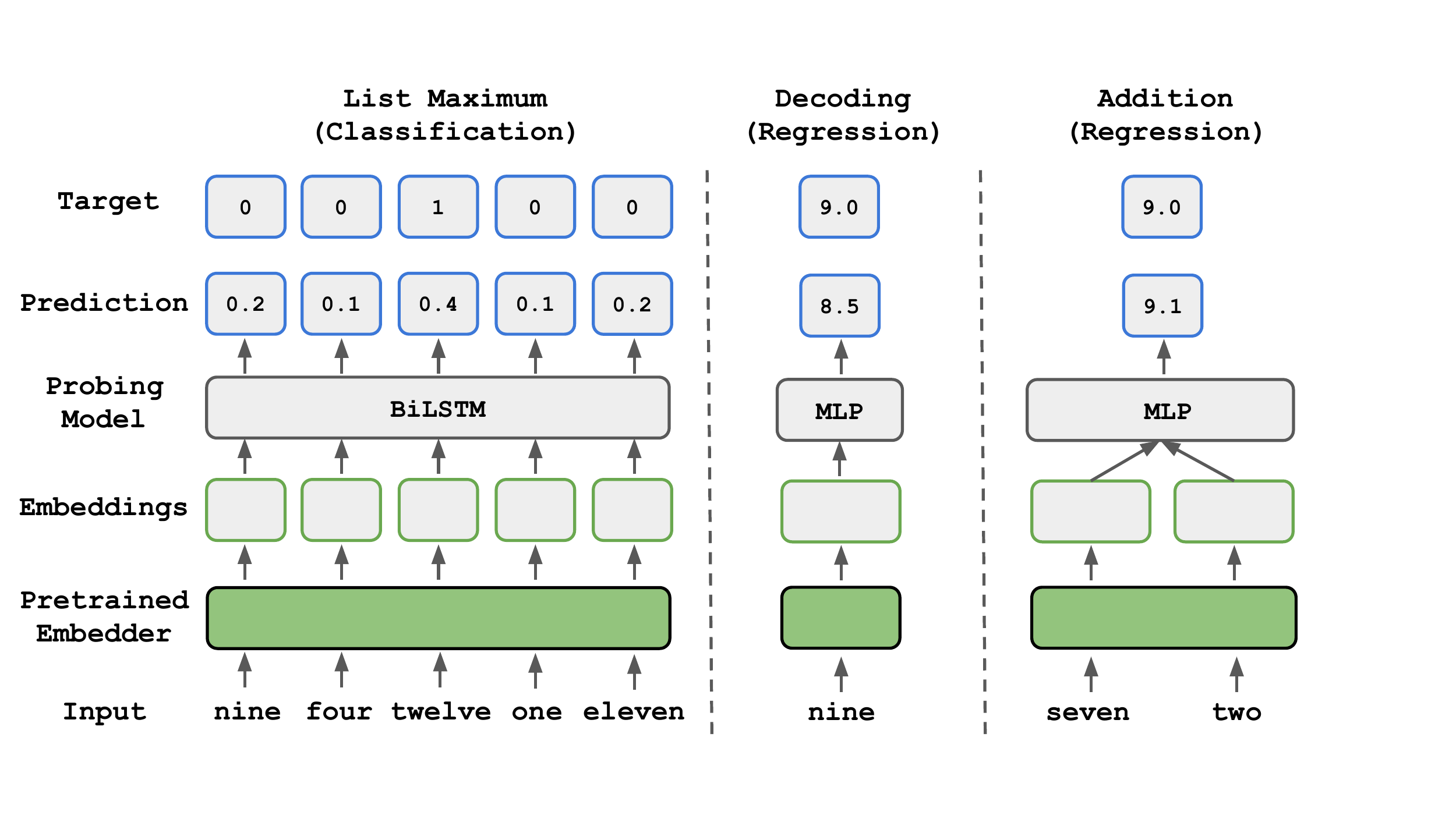}
  \caption{Our probing setup. We pass numbers through a pre-trained embedder (e.g., BERT, GloVe) and train a probing model to solve numerical tasks such as finding a list's maximum, decoding a number, or adding two numbers. If the probing model generalizes to held-out numbers, the pre-trained embeddings must contain numerical information. We provide numbers as either words (shown here), digits (``9''), floats (``9.1''), or negatives (``-9'').}
  \label{fig:probing} 
\end{figure*}

\section{Probing Numeracy of Embeddings}\label{sec:numeracy}

We use synthetic numerical tasks to probe the numeracy of token embeddings. 

\subsection{Probing Tasks}

We consider three synthetic tasks to evaluate numeracy (Figure~\ref{fig:probing}). Appendix~\ref{app:details} provides further details on training and evaluation.

\paragraph{List Maximum} Given a list of the embeddings for five numbers, the task is to predict the index of the maximum number. Each list consists of values of similar magnitude in order to evaluate fine-grained comparisons (see Appendix~\ref{app:details}).
As in typical span selection models~\cite{Seo2017BidirectionalAF}, an LSTM reads the list of token embeddings, and a weight matrix and softmax function assign a probability to each index using the model's hidden state. We use the negative log-likelihood of the maximum number as the loss function.

\paragraph{Decoding} The decoding task probes whether number magnitude is captured (rather than the relative ordering of numbers as in list maximum).  Given a number's embedding, the task is to regress to its value, e.g., the embedding for the string ``five'' has a target of \texttt{5.0}. We consider a linear regression model and a three-layer fully-connected network with ReLU activations. The models are trained using a mean squared error (MSE) loss.

\paragraph{Addition} The addition task requires number manipulation---given the embeddings of two numbers, the task is to predict their sum. Our model concatenates the two token embeddings and feeds the result through a three-layer fully-connected network with ReLU activations, trained using MSE loss. Unlike the decoding task, the model needs to capture number magnitude internally without direct label supervision.

\paragraph{Training and Evaluation} 
We focus on a numerical \emph{interpolation} setting (we revisit extrapolation in Section~\ref{subsec:extrap}): the model is tested on values that are within the training range. We first pick a range (we vary the range in our experiments) and randomly shuffle the integers over it. We then split 80\% of the numbers into a training set and 20\% into a test set. We report the mean and standard deviation across five different random shuffles for a particular range, using the exact same shuffles across all embedding methods.

Numbers are provided as integers (``75''), single-word form (``seventy-five''), floats (``75.1''), or negatives (``-75''). We consider positive numbers less than 100 for word-form numbers to avoid multiple tokens. We report the classification accuracy for the list maximum task (5 classes), and the Root Mean Squared Error (RMSE) for decoding and addition. Note that larger ranges will naturally amplify the RMSE error.
 
\subsection{Embedding Methods}

We evaluate various token embedding methods.~\smallskip

\noindent \textbf{Word Vectors} We use 300-dimensional \glove{}~\cite{pennington2014glove} and \wtovec{} vectors~\cite{mikolov2018advances}. We ensure all values are in-vocabulary for word vectors.~\smallskip

\noindent \textbf{Contextualized Embeddings} We use \elmo{}~\cite{PetersELMo2018} and \bert{}~\cite{devlin2018BERT} embeddings.\footnote{Since our inputs are numbers, not natural sentences, language models may exhibit strange behavior. We experimented with extracting the context-independent feature vector immediately following the character convolutions for \elmo{} but found little difference in results.} \elmo{} uses character-level convolutions of size 1--7 with max pooling. \bert{} represents tokens via sub-word pieces; we use lowercased \bertbase{} with 30k pieces.~\smallskip

\noindent \textbf{\qanet{} Embeddings} We extract the \glove{} embeddings and Char-CNN from the \qanet{} model trained on \drop{}. We also consider an ablation that removes the \glove{} embeddings.

\noindent \textbf{Learned Embeddings} We use a character-level CNN (Char-CNN) and a character-Level LSTM (Char-LSTM). We use left character padding, which greatly improves numeracy for character-level CNNs (details in Appendix~\ref{appendix:padding}).~\smallskip

\noindent \textbf{Untrained Embeddings} We consider two untrained baselines. The first baseline is random token vectors, which trivially fail to generalize (there is no pattern between train and test numbers). These embeddings are useful for measuring the improvement of pre-trained embeddings. We also consider a randomly initialized and untrained Char-CNN and Char-LSTM.~\smallskip

\noindent \textbf{Number's Value as Embedding} The final embedding method is simple: map a number's embedding directly to its value (e.g., ``seventy-five'' embeds to [75]). We found this strategy performs poorly for large ranges; using a base-10 logarithmic scale improves performance. We report this as \emph{Value Embedding} in our results.\footnote{We suspect the failures result from the raw values being too high in magnitude and/or variance for the model. We also experimented with normalizing the values to mean 0 and variance 1; a logarithmic scale performed better.}~\smallskip

All pre-trained embeddings (all methods except the Char-CNN and Char-LSTM) are fixed during training. The probing models are trained on the synthetic tasks on top of these embeddings.

\subsection{Results: Embeddings Capture Numeracy}\label{subsec:natural}

\begin{table*}[t]
    \resizebox{1\linewidth}{!}{
    \begin{tabular}{lccccccccc}
    \toprule
    \bf \textbf{Interpolation} & \multicolumn{3}{c}{\bf List Maximum (5-classes)} & \multicolumn{3}{c}{\bf Decoding (RMSE)} & \multicolumn{3}{c}{\bf Addition (RMSE)} \\
    {\textit{Integer Range}} & [0,99] & [0,999] & [0,9999] & [0,99] & [0,999] & [0,9999] & [0,99] & [0,999] & [0,9999] \\
    \cmidrule(r){1-1}
    \cmidrule(lr){2-4}
    \cmidrule(lr){5-7}
    \cmidrule(lr){8-10}
    Random Vectors & 0.16 & 0.23 & 0.21 & 29.86 & 292.88 & 2882.62 & 42.03 & 410.33 & 4389.39  \\
    Untrained CNN & 0.97 & 0.87 & 0.84 & 2.64 & 9.67 & 44.40 & 1.41 & 14.43 & 69.14\\
    Untrained LSTM & 0.70 & 0.66 & 0.55 & 7.61 & 46.5 & 210.34 & 5.11 & 45.69 & 510.19 \\
    Value Embedding &  \textbf{0.99} & 0.88  & 0.68 & \textbf{1.20} & 11.23 & 275.50 & \textbf{0.30} & 15.98 & 654.33 \\
    \addlinespace
    {\textit{Pre-trained}} & & & & & & \\
    
    Word2Vec &  0.90 &  0.78  &  0.71  & 2.34 & 18.77 & 333.47 & 0.75 & 21.23 & 210.07 \\
    GloVe &  0.90 &  0.78 &  0.72 & 2.23 & 13.77 & 174.21 & 0.80 & 16.51 & 180.31 \\
    ELMo & 0.98 &  0.88 &  0.76 & 2.35 & 13.48 & 62.20 & 0.94 & 15.50 & 45.71\\
    BERT &  0.95 &  0.62 &  0.52 & 3.21 & 29.00 & 431.78 & 4.56 & 67.81 & 454.78 \\
    \addlinespace
    {\textit{Learned}} & & & & & & \\
    Char-CNN & 0.97 &  \textbf{0.93} &  \textbf{0.88} & 2.50 & \textbf{4.92} & \textbf{11.57} & 1.19 & \textbf{7.75} & \textbf{15.09}\\
    Char-LSTM & 0.98 &  0.92 &  0.76 & 2.55 & 8.65 & 18.33 & 1.21 & 15.11 & 25.37\\
    \addlinespace
    {\textit{DROP-trained}} & & & & & & \\
    \qanet{} &  0.91 & 0.81 &  0.72 & 2.99 & 14.19  & 62.17 & 1.11 & 11.33 & 90.01 \\
    \hspace{0.2cm} - GloVe &  0.88 &  0.90 & 0.82 & 2.87  & 5.34 & 35.39 & 1.45 & 9.91 & 60.70 \\
    \bottomrule
    \end{tabular}
    }
    \caption{{\it Interpolation with integers (e.g., ``18'').} All pre-trained embedding methods (e.g., GloVe and ELMo) surprisingly capture numeracy. The probing model is trained on a randomly shuffled 80\% of the \emph{Integer Range} and tested on the remaining 20\%. The probing model architecture and train/test splits are equivalent across all embeddings. We show the mean over 5 random shuffles (standard deviation in Appendix~\ref{appendix:additional}).}
    \label{table:interpolation_digit}
\end{table*}

We find that all pre-trained embeddings \emph{contain} fine-grained information about number magnitude and order. We first focus on integers (Table~\ref{table:interpolation_digit}).

\paragraph{Word Vectors Succeed} Both \wtovec{} and \glove{} significantly outperform the random vector baseline and are among the strongest methods overall. This is particularly surprising given the training methodology for these embeddings, e.g., a continuous bag of words objective can teach fine-grained number magnitude.

\paragraph{Character-level Methods Dominate} Models which use character-level information have a clear advantage over word-level models for encoding numbers. This is reflected in our probing results: character-level CNNs are the best architecture for capturing numeracy. For example, the \qanet{} model without \glove{} (only using its Char-CNN) and \elmo{} (uses a Char-CNN) are the strongest pre-trained methods, and a learned Char-CNN is the strongest method overall. The strength of the character-level convolutions seems to lie in the architectural prior---an \emph{untrained} Char-CNN is surprisingly competitive. Similar results have been shown for images~\cite{saxe2011random}: random CNNs are powerful feature extractors. 

\paragraph{Sub-word Models Struggle} \bert{} struggles for large ranges (e.g., 52\% accuracy for list maximum for [0,9999]). We suspect this results from sub-word pieces being a poor method to encode digits: two numbers which are similar in value can have very different sub-word divisions.

\paragraph{A Linear Subspace Exists} For small ranges on the decoding task (e.g., [0,99]), a linear model is competitive, i.e., a linear subspace captures number magnitude (Appendix~\ref{appendix:additional}). For larger ranges (e.g., [0,999]), the linear model's performance degrades, especially for \bert{}. 

\paragraph{Value Embedding Fails} The Value Embedding method fails for large ranges. This is surprising as the embedding directly provides a number's value, thus, the synthetic tasks \emph{should} be easy to solve. However, we had difficulty training models for large ranges, even when using numerous architecture variants (e.g., tiny networks with 10 hidden units and tanh activations) and hyperparameters. \citet{task2018nalu} discuss similar problems and ameliorate them using new neural architectures.

\paragraph{Words, Floats, and Negatives are Captured} Finally, we probe the embeddings on word-form numbers, floats, and negatives. We observe similar trends for these inputs as integers: pre-trained models exhibit natural numeracy and learned embeddings are strong (Tables~\ref{table:floats}, \ref{table:negative}, and \ref{table:interpolation_words}). The ordering of the different embedding methods according to performance is also relatively consistent across the different input types. One notable exception is that \bert{} struggles on floats, which is likely a result of its sub-word pieces. We do not test \wtovec{} and \glove{} on floats/negatives because they are out-of-vocabulary.

\begin{table}[t]
    \centering
    \begin{tabular}{lcc}
    \toprule
    \bf  \textbf{Interpolation} & \multicolumn{2}{c}{\bf List Maximum (5-classes)}\\
    {\textit{Float Range}} & [0.0,99.9] & [0.0,999.9] \\
    \cmidrule(r){1-1}
    \cmidrule(lr){2-3}
    Rand. Vectors & 0.18 $\pm$ 0.03 & 0.21 $\pm$ 0.04 \\
    ELMo & 0.91 $\pm$ 0.03 & 0.59 $\pm$ 0.01 \\
    BERT & 0.82 $\pm$ 0.05 & 0.51 $\pm$ 0.04\\
    Char-CNN & 0.87 $\pm$ 0.04 & 0.75 $\pm$ 0.03 \\
    Char-LSTM & 0.81 $\pm$ 0.05 & 0.69 $\pm$ 0.02 \\
    \bottomrule
    \end{tabular}    
    \caption{{\it Interpolation with floats (e.g., ``18.1'') for list maximum.} Pre-trained embeddings capture numeracy for float values. The probing model is trained on a randomly shuffled 80\% of the \emph{Float Range} and tested on the remaining 20\%. See the text for details on selecting decimal values. We show the mean alongside the standard deviation over 5 different random shuffles.}
    \label{table:floats}    
\end{table}

\begin{table}[t]
    \centering
    \begin{tabular}{lc}
    \toprule
    \bf  \textbf{Interpolation} & \multicolumn{1}{c}{\bf List Maximum (5-classes)}\\
    {\textit{Integer Range}} & [-50,50]\\
    \cmidrule(r){1-1}
    \cmidrule(lr){2-2}
    Rand. Vectors & 0.23 $\pm$ 0.12 \\
    Word2Vec & 0.89 $\pm$ 0.02\\
    GloVe & 0.89 $\pm$ 0.03 \\
    ELMo & 0.96 $\pm$ 0.01\\
    BERT & 0.94 $\pm$ 0.02\\
    Char-CNN & 0.95 $\pm$ 0.07 \\
    Char-LSTM & 0.97 $\pm$ 0.02 \\
    \bottomrule
    \end{tabular}    
    \caption{{\it Interpolation with negatives (e.g., ``-18'') on list maximum.} Pre-trained embeddings capture numeracy for negative values.}
    \label{table:negative}  
\end{table}

\subsection{Probing Models Struggle to Extrapolate}\label{subsec:extrap}

Thus far, our synthetic experiments evaluate on held-out values \emph{within} the same range as the training data (i.e., numerical interpolation). In Section~\ref{subsec:limit}, we found that \qanet{} struggles to extrapolate to values \emph{outside} the training range. Is this an idiosyncrasy of \qanet{} or is it a more general problem? We investigate this using a numerical extrapolation setting: we train models on a specific integer range and test them on values greater than the largest training number and smaller than the smallest training number.

\setlength{\tabcolsep}{3pt}
\begin{table}[t]
    \centering
    \small
    \begin{tabular}{lccc}
    \toprule
    \bf  \textbf{Extrapolation} & \multicolumn{3}{c}{\bf List Maximum (5-classes)}\\
    {\textit{Test Range}} & [151,160] & [151,180] & [151,200]\\
    \cmidrule(r){1-1}
    \cmidrule(lr){2-4}
    Rand. Vectors & 0.17 & 0.22 & 0.15 \\
    Untrained CNN & 0.80 & 0.47 & 0.41\\
    \addlinespace
    {\textit{Pre-trained}} & & &\\
    Word2Vec & 0.14 & 0.16 & 0.11 \\
    GloVe & 0.19 & 0.17 & 0.21 \\
    ELMo & 0.65 & 0.57 & 0.38\\
    BERT & 0.35 & 0.11 & 0.14\\
    \addlinespace
    {\textit{Learned}} & & &\\
    Char-CNN & 0.81 & 0.75 & 0.73\\
    Char-LSTM & \textbf{0.88} & \textbf{0.84} & \textbf{0.82}\\
    \addlinespace
    {\textit{DROP}} & & & \\
    \qanet{} & 0.31 & 0.29 & 0.25\\
    \hspace{0.2cm} - GloVe & 0.58 & 0.53 & 0.48\\
    \bottomrule
    \end{tabular}    
    \caption{{\it Extrapolation on list maximum.} The probing model is trained on the integer range [0,150] and evaluated on integers from the \emph{Test Range}. The probing model struggles to extrapolate when trained on the pre-trained embeddings.}
    \label{table:extrapolation_digits}
\end{table}

\paragraph{Extrapolation for Decoding and Addition} For decoding and addition, models struggle to extrapolate. Figure~\ref{fig:plot} shows the predictions for models trained on 80\% of the values from [-500,500] and tested on held-out numbers in the range [-2000, 2000] for six embedding types. The embedding methods fail to extrapolate in different ways, e.g., predictions using \wtovec{} decrease almost monotonically as the input increases, while predictions using \bert{} are usually near the highest training value. \citet{task2018nalu} also observe that models struggle outside the training range; they attribute this to failures in neural models themselves.

\paragraph{Extrapolation for List Maximum} For the list maximum task, accuracies are closer to those in the interpolation setting, however, they still fall short. Table~\ref{table:extrapolation_digits} shows the accuracy for models trained on the integer range [0,150] and tested on the ranges [151,160], [151,180], and [151,200]; all methods struggle, especially token vectors. 

\paragraph{Augmenting Data to Aid Extrapolation} Of course, in many real-word tasks it is possible to ameliorate these extrapolation failures by augmenting the training data (i.e., turn extrapolation into interpolation). Here, we apply this idea to aid in training \qanet{} for \drop{}. For each superlative and comparative example, we duplicate the example and modify the numbers in its paragraph using the \emph{Add} and \emph{Multiply} techniques described in Section \ref{subsec:limit}. Table \ref{table:count} shows that this data augmentation can improve both interpolation and extrapolation, e.g., the accuracy on superlative questions with large numbers can \emph{double}. 
\newpage
\section{Discussion and Related Work}\label{sec:related}

An open question is how the training process elicits numeracy for word vectors and contextualized embeddings. Understanding this, perhaps by tracing numeracy back to the training data, is a fruitful direction to explore further (c.f., influence functions~\cite{koh2017influence,brunet2019understanding}).

More generally, numeracy is one type of emergent knowledge. For instance, embeddings may capture the size of objects~\cite{forbes2017verb}, speed of vehicles, and many other ``commonsense'' phenomena~\cite{yang2018common}. \citet{vendrov2015order} introduce methods to encode the order of such phenomena into embeddings for concepts such as hypernymy; our work and \citet{yang2018common} show that a relative ordering naturally emerges for certain concepts.

In concurrent work, \citet{naik2019numeracy} also explore numeracy in word vectors. Their methodology is based on variants of nearest neighbors and cosine distance; we use neural network probing classifiers which can capture highly non-linear dependencies between embeddings. We also explore more powerful embedding methods such as ELMo, BERT, and learned embedding methods.

\paragraph{Probing Models} Our probes of \emph{numeracy} parallel work in understanding the linguistic capabilities (\emph{literacy}) of neural models~\cite{conneau2018cram,liu2019linguistic}. LSTMs can remember sentence length, word order, and which words were present in a sentence~\cite{adi2017lstm}. \citet{khandelwal2018lm} show how language models leverage context, while \citet{linzen2016assessing} demonstrate that language models understand subject-verb agreement.

\paragraph{Numerical Value Prediction} \citet{Spithourakis2018numeracy} improve the ability of language models to \emph{predict} numbers, i.e., they go beyond categorical predictions over a fixed-size vocabulary. They focus on improving models; our focus is probing embeddings. \citet{kotnis2019learning} predict numerical attributes in knowledge bases, e.g., they develop models that try to predict the population of Paris.

\paragraph{Synthetic Numerical Tasks} Similar to our synthetic numerical reasoning tasks, other work considers sorting~\cite{graves2014neural}, counting~\cite{weiss2018counting}, or decoding tasks~\cite{task2018nalu}. They use synthetic tasks as a testbed to prove or design better models, whereas we use synthetic tasks as a probe to understand token embeddings. In developing the Neural Arithmetic Logic Unit, \citet{task2018nalu} arrive at similar conclusions regarding extrapolation: neural models have difficulty outputting numerical values outside the training range.
\section{Conclusion}\label{sec:conclusion}

How much do NLP models know about numbers? By digging into a surprisingly successful model on a numerical reasoning dataset (DROP), we discover that pre-trained token representations naturally encode numeracy.

We analyze the limits of this numeracy, finding that CNNs are a particularly good prior (and likely the cause of ELMo's superior numeracy compared to BERT) and that it is difficult for neural models to extrapolate beyond the values seen during training. There are still many fruitful areas for future research, including discovering \emph{why} numeracy naturally emerges in embeddings, and what other properties are similarly emergent. 
\section*{Acknowledgements}

We thank Mark Neumann, Suchin Gururangan, Pranav Goel, 
Shi Feng, Nikhil Kandpal, Dheeru Dua, the members of AllenNLP and UCI NLP, and 
the reviewers for their valuable feedback.

\bibliography{journal-abbrv,bib}
\bibliographystyle{acl_natbib}

\appendix
\clearpage
\section{\qanet{} Details and Numeracy Requirements}\label{appendix:qanet}

Aside from typical paragraph spans, \qanet{} can also output spans in the question, counts from 0--9, and arithmetic expressions. The arithmetic expressions are generated by first extracting all the numbers from the passage and then assigning a coefficient in \{-1,0,1\} to each number. The final answer is calculated as the sum of each number using its associated coefficient.

Addition and subtraction questions are present in \drop{} but do not require numeracy for \qanet{}: the model can output the correct coefficients based on the surrounding context words without understanding number magnitude. For example, consider a paragraph containing,  ``Superbowl XXXI occurred in 1997 and Superbowl XXXVII occurred in 2003'', and a question ``How many years after Superbowl XXXI did Superbowl XXXVII occur?''. The model can output coefficient +1 on ``2003'' and -1 on ``1997'' and answer correctly without understanding the magnitude of the two years. Counting questions also do not require reading or manipulating numbers in text: the model only needs to output the correct 10-way classification value. 

\section{Left-padding for Enhancing Char-CNN Numeracy}\label{appendix:padding}

In modern deep learning and NLP frameworks, pad tokens are often appended to the right of inputs to ensure equal lengths inside each batch (for efficiency). When using \emph{character}-level convolutions, padding must be added to the characters of a word when its length is smaller than the minimum kernel width (for correctness). Naturally, the right padding implementation is re-used, e.g., the two numbers ``11'' and ``110'' are represented as:
\begin{align}
\begin{array}{rcl}
\nonumber
 11 &=& 11\oldemptyset\oldemptyset\oldemptyset\\
 110 &=& 110\oldemptyset\oldemptyset,
\end{array}
\label{eq:mape}
\end{align}
where $\oldemptyset$ is the pad token. However, right padding is detrimental to learning numeracy: the ``hundreds'' and ``tens'' digits are now in the same position in the two tokens. We found that left padding numbers substantially decreases the generalization error and convergence rate for Char-CNN models trained on the synthetic tasks.\footnote{Padding on the left and right via {\tt SAME} convolutions also mitigates this issue.}

\section{Training Details for Probing}\label{app:details}
 
We create training/test splits for the addition task in the following manner. We first shuffle and split an integer range, putting 80\% into train and 20\% into test. Next, we enumerate all possible pairs of numbers in the two respective splits. When using large ranges such as [0,999], we sub-sample a random 10\% of the training and test pairs. 

For the list maximum task, we first shuffle and split the data, putting 80\% into a training pool of numbers and 20\% into a test pool. In initial experiments, we created the lists of five numbers by sampling uniformly over the training/test pool. However, as the random samples will likely be spread out over the range, the numbers are easy to distinguish. We instead create 100,000 training examples and 10,000 examples in the following manner. We first sample a random integer from the training or test pool. Next, we sample from a Gaussian with mean zero and variance equal to 0.01 times the total size of the range. Finally, we add the random Gaussian sample to the random integer, and round to the nearest value in the pool. This forces the numbers to be nearby.

\section{Additional Synthetic Results}\label{appendix:additional}

\subsection{Linear Regression Accuracies}

Table~\ref{table:linear_regression} provides accuracies for number decoding using linear regression (interpolation setting).

\begin{table}[h!]
    \centering
    \begin{tabular}{lccc}
    \toprule
    \bf Interpolation & \multicolumn{3}{c}{\bf Decoding (RMSE)}  \\
    {\textit{Integer Range}} & [0,50] & [-50,50] & [0,999] \\
    \cmidrule(r){1-1}
    \cmidrule(lr){2-4}
    Random & 13.86 & 29.46 & 275.41 \\
    Word2Vec & 4.15 & 8.93 & 29.04\\
    GloVe & 3.21 & 5.76 & 23.27\\
    ELMo & 1.20 & 2.89 & 21.53 \\
    BERT & 3.23 & 7.86 & 64.42\\
    \bottomrule
    \end{tabular}
    \caption{{\it Number Decoding interpolation accuracy with linear regression.} Linear regression is competitive to the fully connected probe for smaller numbers.}
    \label{table:linear_regression}
\end{table}

\subsection{Variances on Digit Form Results}

Table~\ref{table:interpolation_digit_variance} shows the mean and standard deviation for the synthetic tasks using five random shuffles.

\newpage
\subsection{Float Values}

We test floats with one decimal point. We follow the setup for the list maximum task (Appendix~\ref{app:details}) with a minor modification. For 50\% of the training/test lists, we reuse the same integer five times but sample a different random value to put after the decimal point. For example, 50\% of the lists are of the form: [15.3, 15.6, 15.1, 15.8, 15.2] (the same base integer is repeated with a different decimal), and 50\% are random integers with a random digit after the decimal: [11.7, 16.4, 9.3, 7.9, 13.3]. This forces the model to consider the numbers on both the left and the right of the decimal. 

\subsection{Word Form Results}

Table~\ref{table:interpolation_words} presents the results using word-forms. We do not use numbers larger than 100 as they consist of multiple words. 

\section{Automatically Modifying DROP Paragraphs}\label{appendix:internal}

Modifying a \drop{} paragraph automatically is challenging as it may change the answer to the question if done incorrectly. We also cannot modify the answer because many \drop{} questions have count answers. To guarantee that the original annotated answer can still be used, we perform the number transformation in the following manner. We keep the original passage text the same, but, we modify the model's internal embeddings for the numbers directly. In other words, the model uses exactly the same embeddings for the original text except for the modified numbers. The model then needs to find the correct index (e.g. the index of the correct span, or the index of the correct number) given these modified embeddings.

\begin{table*}[t]
    \resizebox{1\linewidth}{!}{
    \begin{tabular}{lccccccccc}
    \toprule
    \bf \textbf{Interpolation} & \multicolumn{3}{c}{\bf List Maximum (5-classes)} & \multicolumn{3}{c}{\bf Decoding (RMSE)}  & \multicolumn{3}{c}{\bf Addition (RMSE)} \\
    {\textit{Integer Range}} & [0,99] & [0,999] & [0,9999] & [0,99] & [0,999] & [0,9999]  & [0,99] & [0,999] & [0,9999] \\
    \cmidrule(r){1-1}
    \cmidrule(lr){2-4}
    \cmidrule(lr){5-7}
    \cmidrule(lr){8-10}
    Random Vectors & 0.16 $\pm$ 0.03 &  0.23 $\pm$ 0.12 & 0.21 $\pm$ 0.02 & 29.86 $\pm$ 4.44 & 292.88 $\pm$ 13.48 & 2882.62 $\pm$ 71.68 & 42.03 $\pm$ 7.79 & 410.33 $\pm$ 16.05 & 4389.39 $\pm$ 310.91 \\
    Untrained CNN & 0.97 $\pm$ 0.01 & 0.87 $\pm$ 0.02 &  0.84 $\pm$ 0.03 & 2.64 $\pm$ 0.68 & 9.67 $\pm$ 1.17 & 44.40 $\pm$ 4.98 & 1.41 $\pm$ 0.05 & 14.43 $\pm$ 1.90 & 69.14 $\pm$ 21.54 \\
    Untrained LSTM & 0.70 $\pm$ 0.05 & 0.66 $\pm$ 0.03 & 0.55 $\pm$ 0.02 & 7.61 $\pm$ 1.33 & 46.5 $\pm$ 5.65 & 210.34 $\pm$ 9.91 & 5.11 $\pm$ 2.1 & 45.69 $\pm$ 3.78 & 510.19 $\pm$ 31.45 \\
    Value Embedding &  0.99 $\pm$ 0.01 & 0.88 $\pm$ 0.04 & 0.68 $\pm$ 0.06 & 1.20 $\pm$ 0.76  & 11.23 $\pm$ 1.35 & 275.5 $\pm$ 135.59 & 0.30 $\pm$ 0.01 & 15.98 $\pm$ 3.62 & 654.33 $\pm$ 69.91 \\
    \addlinespace
    {\textit{Pre-trained}} & & & & & & \\
    
    Word2Vec &  0.90 $\pm$ 0.03 &  0.78 $\pm$ 0.05 &  0.71 $\pm$ 0.03  & 2.34 $\pm$ 1.44 & 18.77 $\pm$ 4.40 & 333.47 $\pm$ 14.83 & 0.75 $\pm$ 0.41 & 21.23 $\pm$ 3.53 & 210.07 $\pm$ 30.56 \\
    GloVe &  0.90 $\pm$ 0.02 &  0.78 $\pm$ 0.04 &  0.72 $\pm$ 0.02 & 2.23 $\pm$ 1.26 & 13.77 $\pm$ 3.23 & 174.21 $\pm$ 31.91 & 0.80 $\pm$ 0.30 & 16.51 $\pm$ 1.17 & 180.31 $\pm$ 18.97 \\
    ELMo & 0.98 $\pm$ 0.04 &  0.88 $\pm$ 0.02 &  0.76 $\pm$ 0.04 & 2.35 $\pm$ 0.65 & 13.48 $\pm$ 2.19 & 62.2 $\pm$ 5.04 & 0.94 $\pm$ 0.35 & 15.50 $\pm$ 2.49 & 45.71 $\pm$ 18.69  \\
    BERT &  0.95 $\pm$ 0.02 &  0.62 $\pm$ 0.01 &  0.52 $\pm$ 0.07 & 3.21 $\pm$ 0.31 & 29.00 $\pm$ 7.93 & 431.78 $\pm$ 30.27 & 4.56 $\pm$ 1.76 & 67.81 $\pm$ 30.34 & 454.78 $\pm$ 91.07\\
    \addlinespace
    {\textit{Learned}} & & & & & & \\
    Char-CNN & 0.97 $\pm$ 0.02 &  0.93 $\pm$ 0.01 & 0.88 $\pm$ 0.02 & 2.50 $\pm$ 0.49 & 4.92 $\pm$ 0.97 & 11.57 $\pm$ 1.34 & 1.19 $\pm$ 0.18 & 7.75 $\pm$ 1.85 & 15.09 $\pm$ 0.90 \\
    Char-LSTM & 0.98 $\pm$ 0.02 &  0.92 $\pm$ 0.02 &  0.76 $\pm$ 0.03 & 2.55 $\pm$ 1.81 & 8.65 $\pm$ 1.11 & 18.33 $\pm$ 3.21 & 1.21 $\pm$ 0.10 & 15.11 $\pm$ 1.87 & 25.37 $\pm$ 3.40 \\
    \addlinespace
    {\textit{DROP-trained}} & & & & & & \\
    \qanet{} &  0.91 $\pm$ 0.01 & 0.81 $\pm$ 0.02 &  0.72 $\pm$ 0.03 & 2.99 $\pm$ 1.11 & 14.19 $\pm$ 3.75 & 62.17 $\pm$ 4.31 & 1.11 $\pm$ 0.41 & 11.33 $\pm$ 1.67 & 90.01 $\pm$ 17.88 \\
    \hspace{0.2cm} - GloVe &  0.88 $\pm$ 0.02 & 0.90 $\pm$ 0.03 & 0.82 $\pm$ 0.02 &
    2.87 $\pm$ 0.83 & 5.34 $\pm$ 0.77 & 35.39 $\pm$ 4.32 & 1.45 $\pm$ 0.95 & 9.91 $\pm$ 1.45 & 60.70 $\pm$ 13.04 \\
    \bottomrule
    \end{tabular}
    }
    \caption{Mean and standard deviation for Table~\ref{table:interpolation_digit} (interpolation tasks with integers).}
    \label{table:interpolation_digit_variance}
\end{table*}

\begin{table*}[t]
    \resizebox{1\linewidth}{!}{
    \begin{tabular}{lcccccc}
    \toprule
    \bf \textbf{Interpolation} & 
    \multicolumn{2}{c}{\bf List Maximum (5-classes)}
    & \multicolumn{2}{c}{\bf Decoding (MSE)} & \multicolumn{2}{c}{\bf Addition (MSE)}  \\
    {\textit{Integer Range}} & [zero, fifty] & [zero,ninety-nine] & [zero, fifty] & [zero,ninety-nine] & [zero, fifty] & [zero,ninety-nine] \\
    \cmidrule(r){1-1}
    \cmidrule(lr){2-3}
    \cmidrule(lr){4-5}
    \cmidrule(lr){6-7}
    Random Vectors & 0.16 $\pm$ 0.03 & 0.23 $\pm$ 0.12 & 15.43 $\pm$ 1.13  & 29.86 $\pm$ 4.44 & 23.14 $\pm$ 4.46 & 43.86 $\pm$ 1.41 \\
    \addlinespace
    {\textit{Pre-trained}} & & & & & & \\
    Word2Vec & 0.96 $\pm$ 0.05 & 0.91 $\pm$ 0.05 & 2.40 $\pm$ 0.64 & 3.94 $\pm$ 1.88 & 2.10 $\pm$ 0.31 & 4.10 $\pm$ 0.98 \\
    GloVe  & 0.93 $\pm$ 0.07 & 0.90 $\pm$ 0.03 & 3.02 $\pm$ 0.98 & 4.75 $\pm$ 1.59 & 3.33 $\pm$ 0.51 & 4.81 $\pm$ 1.00 \\
    ELMo & 0.75 $\pm$ 0.04 & 0.82 $\pm$ 0.10 & 3.86 $\pm$ 0.92 & 7.31 $\pm$ 2.06 & 3.91 $\pm$ 0.55 & 6.97 $\pm$ 1.19  \\
    BERT & 0.61 $\pm$ 0.11 & 0.66 $\pm$ 0.05 & 7.59 $\pm$ 1.37 & 13.55 $\pm$ 3.15 & 7.33 $\pm$ 1.05 & 12.91 $\pm$ 2.57 \\
    \addlinespace
    {\textit{Learned}} & & & & & & \\
    Char-CNN & 0.98 $\pm$ 0.01 & 0.99 $\pm$ 0.01 & 5.73 $\pm$ 0.93 & 7.85 $\pm$ 2.36 & 6.11 $\pm$ 0.55 & 7.11 $\pm$ 2.34 \\
    Char-LSTM & 0.89 $\pm$ 0.06 & 0.86 $\pm$ 0.08 & 8.45 $\pm$ 4.97 & 7.31 $\pm$ 2.06 & 8.91 $\pm$ 3.11 & 9.51 $\pm$ 1.19 \\
    \addlinespace
    {\textit{DROP-trained}} & & & & & & \\
    \qanet{} & 0.98 $\pm$ 0.01 & 0.97 $\pm$ 0.02 & 3.17 $\pm$ 1.05 & 4.31 $\pm$ 0.68 & 3.05 $\pm$ 0.87 & 3.95 $\pm$ 0.33 \\
    \hspace{0.5cm} - GloVe & 0.96 $\pm$ 0.03 & 0.97 $\pm$ 0.02 & 7.81 $\pm$ 1.34 & 9.43 $\pm$ 2.31 & 8.55 $\pm$ 0.88 & 10.01 $\pm$ 2.07\\
    \bottomrule
    \end{tabular}
    }
    \caption{{\it Interpolation task accuracies with word form (e.g., ``twenty-five'').} The model is trained on a randomly shuffled 80\% of the \emph{Integer Range} and tested on the remaining 20\%. We show the mean and standard deviation for five random shuffles.}
    \label{table:interpolation_words}
\end{table*}

\section{Extrapolation with Data Augmentation}

For each superlative and comparative example, we modify the numbers in its paragraph using the \emph{Add} and \emph{Multiply} techniques mentioned in Section \ref{subsec:limit}. We first multiply the paragraph's numbers by a random integer from [1, 10], and then add another random integer from [0, 20]. We train \qanet{} on the original paragraph and an additional modified version for all training examples. We use a single additional paragraph for computational efficiency; augmenting the data with more modified paragraphs may further improve results.

We test \qanet{} on the original validation set, as well as a \emph{Bigger} validation set. We created the Bigger validation set  by multiplying each paragraph's numbers by a random integer from [11,20] and then adding a random value from [21,40]. Note that this range is larger than the one used for data augmentation. 

Table \ref{table:count} shows the results of \qanet{} trained with data augmentation. Data augmentation provide small gains on the original superlative and comparative question subset, and significant improvements on the \emph{Bigger} version (it doubles the model's F1 score for superlative questions). 

\begin{table*}[t]
    \centering \small
    \begin{tabular}{lcccccc}
    \toprule
    & \multicolumn{2}{c}{\textbf{Superlative}} & \multicolumn{2}{c}{\textbf{Comparative}} & \multicolumn{2}{c}{\textbf{All Validation}} \\
    & Original & Bigger & Original & Bigger & Original & Bigger \\
    \cmidrule(lr){2-3}
    \cmidrule(lr){4-5}
    \cmidrule(lr){6-7}
    NAQANet & 64.5 / 67.7 &  30.0 / 32.2 & 73.6 / 76.4 & 70.3 / 73.0 &  \textbf{46.2} / 49.2 & 38.7 / 41.4  \\
    + Data Augmentation & \textbf{67.6} / \textbf{70.9} & \textbf{59.2} / \textbf{62.4} & \textbf{76.0} / \textbf{77.7} & \textbf{75.0} / \textbf{76.8} & 46.1 / \textbf{49.3} & \textbf{42.8} /  \textbf{45.8} \\
    \bottomrule
    \end{tabular}
    \caption{Data augmentation improves \qanet{}'s interpolation and extrapolation results. We created the \emph{Bigger} version of \drop{} by multiplying numbers in the passage by a random integer from [11, 20] and then adding a random integer from [21, 40]. Scores are shown in EM / F1 format.}
    \label{table:count}
\end{table*}

\end{document}